\documentclass[mlabstract]{jmlr}

\usepackage{longtable}

\usepackage{booktabs}
\usepackage[load-configurations=version-1]{siunitx} 

\usepackage{wrapfig}

\theorembodyfont{\upshape}
\theoremheaderfont{\scshape}
\theorempostheader{:}
\theoremsep{\newline}

\jmlrvolume{}
\firstpageno{1}
\editors{List of editors' names}

\jmlryear{2025}
\jmlrworkshop{Symmetry and Geometry in Neural Representations}

\title[Beyond I-Con]{Beyond I-Con: A Roadmap for Representation Learning Loss Discovery}

  \author{\Name{Jasmine Shone}\thanks{Equal contribution} \Email{jasshone@mit.edu}\\
  \Name{Zhening Li}\footnotemark[1] \Email{zli11010@csail.mit.edu}\\
  \Name{Shaden Alshammari} \Email{shaden@mit.edu}\\
  \Name{Mark Hamilton} \Email{marhamil@mit.edu}\\
  \Name{William Freeman} \Email{billf@mit.edu}\\
  \addr MIT CSAIL, Cambridge, MA}

\begin{document}

\maketitle

\begin{abstract}
The Information Contrastive (I-Con) framework revealed that over 23 representation learning methods implicitly minimize KL divergence between data and learned distributions that encode similarities between data points. However, a KL-based loss may be misaligned with the true objective, and properties of KL divergence such as asymmetry and unboundedness may create optimization challenges. We present Beyond I-Con, a framework that enables systematic discovery of novel loss functions by exploring alternative statistical divergences. Key findings: (1) on unsupervised clustering of DINO-ViT embeddings, we achieve state-of-the-art results by modifying the PMI algorithm to use total variation (TV) distance; (2) supervised contrastive learning with Euclidean distance as the feature space metric is improved by replacing the standard loss function with Jenson-Shannon divergence (JSD); (3) on dimensionality reduction, we achieve superior qualitative results and better performance on downstream tasks than SNE by replacing KL with a bounded $f$-divergence. Our results highlight the importance of considering divergence choices in representation learning optimization.
\end{abstract}

\section{Introduction}

The choice of optimization objective fundamentally determines the success of representation learning methods, yet the field has largely focused on a single statistical divergence measure without systematic exploration of alternatives. The Information Contrastive (I-Con) framework recently revealed that over 23 diverse representation learning methods all implicitly minimize KL divergence between data and learned distributions that encode similarities between data points \citep{alshammari2024icon}. The natural question emerges: if representation learning methods can be unified under minimizing the divergence between two distributions, what happens when we systematically explore alternative divergences?

\textbf{Contributions.} We present Beyond I-Con, making the following contributions: (1) We generalize I-Con by replacing KL divergence with alternative $f$-divergences, revealing that KL is not unique in enabling meaningful feature optimization; (2) We systematically explore $f$-divergences, uncovering novel loss functions with superior performance on unsupervised clustering, supervised contrastive learning, and dimensionality reduction.

\section{Background: I-Con Overview}
The I-Con framework unifies representation learning methods by framing them as ``minimizing the average KL divergence between two conditional `neighborhood distributions' that define transition probabilities between data points,'' which we index with $i = 1, \ldots, N$ \citep{alshammari2024icon}. $p(j \mid i)$ is typically a fixed ``supervisory'' distribution, and $q_{\phi}(j \mid i)$ are learnable transition probabilities typically calculated using similarities between features --- see Figure 2a of the original paper for an illustration. The core I-Con loss function (Equation 1 of the original paper) is
\begin{equation}
\mathcal{L}_{\text{I-Con}} = \mathbb{E}_{i \sim p(i)} \left[ \mathcal{D}_\text{KL}\big(p(\cdot|i) \big\Vert q(\cdot|i)\big) \right].
\end{equation}
By varying how $p$ is constructed from the dataset and how $q$ is defined in terms of similarities between features, $\mathcal{L}_{\text{I-Con}}$ reproduces the loss functions of many existing representation learning methods.

\section{Beyond I-Con Framework}

We generalize the I-Con objective by replacing the KL divergence with any positive-definite divergence $\mathcal{D}$:
\begin{equation}
\mathcal{L}_{\text{Beyond I-Con}} = \mathbb{E}_{i \sim p(i)} \left[ \mathcal{D}\big(p(\cdot|i) \big\Vert q(\cdot|i)\big) \right]
\end{equation}

We focus on \textbf{$f$-divergences} including KL, Total Variation (TV), Jensen-Shannon (JSD), and Hellinger because they are most directly comparable to KL as a measure of distance between distributions --- some losses such as JSD directly remedy weaknesses of KL such as asymmetry and unboundedness.








\vspace{-10pt}
\section{Experimental Results}

\subsection{Unsupervised Clustering}
We modify the Pointwise Mutual Information (PMI) clustering algorithm \citep{adaloglou2023exploringlimitsdeepimage} by using different divergences than KL. We followed the same training setup as in the I-Con paper, which clusters DINO ViT embeddings \citep{caron2021emerging} on ImageNet-1K \citep{deng2009imagenet} by training a linear classifier for 30 epochs with a batch size of 4096, a learning rate of $1 \times 10^{-3}$, and the Adam optimizer \citep{kingma2014adam}. Data augmentation was applied to the training samples. Table 1 contains the results of this experiment. Clustering using PMI with TV outperforms the state-of-the-art on ViT-B/14 and ViT-L/14 embeddings.

\begin{table}[ht]
\centering
\small
\begin{tabular}{lccc}
\toprule
Method & DiNO ViT-S/14 & DiNO ViT-B/14 & DiNO ViT-L/14 \\
\midrule
k-Means & 51.84 & 52.26 & 53.36 \\
TEMI \citep{adaloglou2023exploringlimitsdeepimage} & 56.84 & 58.62 & --- \\
Debiased InfoNCE Clustering & \textbf{57.8 $\pm$ 0.26} & 64.75 $\pm$ 0.18 & 67.52 $\pm$ 0.28 \\
\midrule
JSD & 53.50 & 63.80  & 66.60\\
TV & 55.90 & \textbf{65.13 $\pm$ 0.13} & \textbf{68.40 $\pm$ 0.29} \\
Hellinger & 54.90 & 63.80 & 67.85\\
\bottomrule
\label{tab:clusterin}
\end{tabular}
\caption{Comparison of methods on ImageNet-1K clustering with respect to Hungarian Accuracy. TV outperforms the state-of-the-art for ViT-B and ViT-L.}
\end{table}
\vspace{-10pt}

\subsection{Supervised Contrastive Learning}

\begin{table}
\centering
\small
\begin{tabular}{lcc}
\toprule
Divergence & Linear probe test acc. & $k$-NN ($k = 7$) test acc. \\
\midrule
KL & 90.03 $\pm$ 0.14 & 89.61 $\pm$ 0.13 \\
\midrule
TV & 83.23 $\pm$ 0.18 & 82.95 $\pm$ 0.16 \\
Hellinger & 90.47 $\pm$ 0.08 & 90.40 $\pm$ 0.09 \\
JSD & \textbf{90.84 $\pm$ 0.11} & \textbf{90.62 $\pm$ 0.11} \\
\bottomrule
\end{tabular} \\
\vspace{5pt}
\caption{Downstream classification accuracy from SupCon-learned features on CIFAR-10. Errors are standard errors of the mean over 5 seeds.}
\label{tab:supcon}
\end{table}

We trained ResNet-50 \citep{he2016deep} models with supervised contrastive learning \citep{khosla2020supervised} on CIFAR-10 \citep{krizhevsky2009learning}.
We used a Euclidean distance metric on features
and trained for 150 epochs with a batch size of 2048 and learning rate of 1e-3.
We systematically varied divergence measures.
We used the learned features to perform classification by training a linear probe or applying $k$-nearest neighbors ($k$-NN). Results are shown in Table \ref{tab:supcon}.

Both Hellinger distance and Jenson-Shannon divergence (JSD) exhibit better performance than vanilla supervised contrastive learning (KL divergence).
In particular, JSD achieves the best performance.


\subsection{Dimensionality Reduction}

\begin{figure}[t]
\centering
\includegraphics[width=1\textwidth]{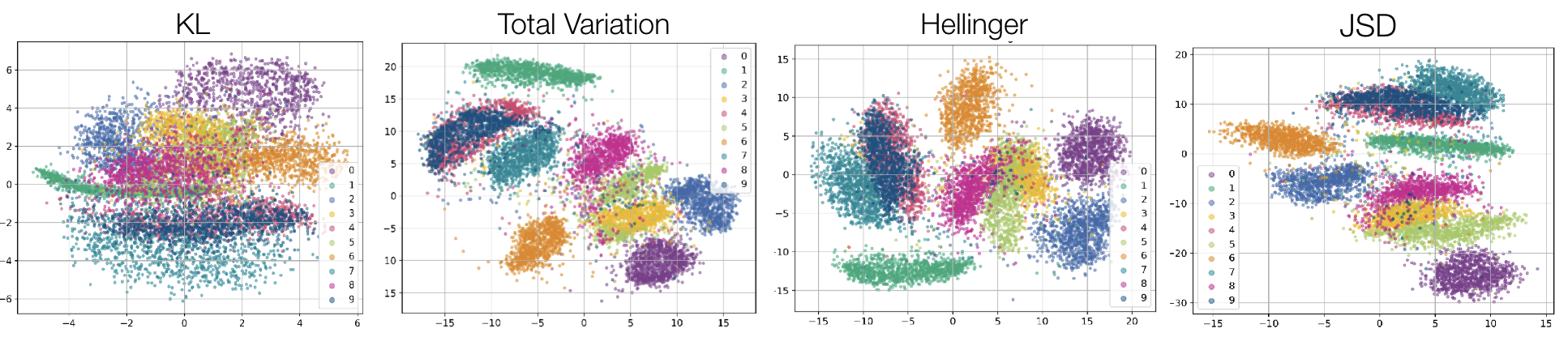}
\caption{Results for running SNE on CIFAR-10 using different divergences, after 150 epochs with a CNN model architecture at learning rate 1e-3. Each color represents a class. KL divergence produces highly overlapping categories in the SNE visualization while other divergences achieve separation.}
\label{fig:CIFAR10}
\end{figure}

We also ran SNE \citep{hinton2002sne} with a CNN backbone on CIFAR-10
using different divergences to visualize qualitative differences between resulting image embeddings, as shown in Figure \ref{fig:CIFAR10}.
Visually, while SNE with KL divergence creates highly overlapping clusters,
the clusters resulting from SNE with the other divergences are more cleanly separated.

\section{Analysis and Discussion}
Across the tasks of unsupervised clustering, supervised contrastive learning, and dimensionality reduction, we observe that a non-KL loss outperforms KL. 
We hypothesize this is because optimizing for KL overly penalizes placing dissimilar points farther apart in the feature space (KL diverges to infinity as $q(j \mid i) \to 0$).
This causes different clusters or classes to crowd together and start overlapping with each other.

For example, in dimensionality reduction,
SNE \citep{hinton2002sne} is known to have this crowding problem, which we also observe in Figure \ref{fig:CIFAR10}.
Beyond I-Con provides a solution
by replacing the KL divergence in the loss with other distances/divergences.
TV, JSD, and Hellinger all penalize small values of $q(j \mid i)$ less than KL does,
since they all remain bounded as $q(j \mid i) \to 0$.\footnote{
t-SNE \citep{van2008visualizing} provides an alternative solution to the crowding problem by replacing $q(\cdot \mid i)$
with one where $q(j \mid i)$ does not decrease as quickly as the distance between
$x_i$ and $x_j$ increases.}
Indeed, they all solve the crowding problem as seen in the resultant
low-dimensional visualizations (Figure \ref{fig:CIFAR10}).
Appendix \ref{apd:geometry} provides further analysis on the geometric arrangement of clusters
in high- vs.\ low-dimensional spaces.

We also hypothesize that KL-based losses may be prone to unstable gradients during training, similar to previous findings \citep{lazic2021,wassersteingan}. For example, in dimensionality reduction, our gradient norm plots during training (Figure \ref{fig:norm}) show large spikes in gradients near the beginning of training when the KL-based loss from vanilla SNE is used. 




\section{Conclusion}
In this work, we extended the I-Con representation learning framework with a new dimension --- the type of $f$-divergence in the loss function.
By experimenting with alternative divergences,
we achieve state-of-the-art ImageNet-1K clustering, surpass vanilla supervised contrastive learning on CIFAR-10, and outperform SNE on CIFAR-10.
Beyond I-Con challenges the default reliance on KL divergence in representation learning by showing that alternative statistical divergences can yield superior performance and serve as a basis for novel loss discovery.

\label{sec:cite}

\bibliography{references}

@inproceedings{
  alshammari2024icon,
  title={A Unifying Framework for Representation Learning},
  author={Shaden Naif Alshammari and Mark Hamilton and Axel Feldmann and John R. Hershey and William T. Freeman},
  booktitle={The Thirteenth International Conference on Learning Representations},
  year={2025},
  url={https://openreview.net/forum?id=WfaQrKCr4X}
}

@inproceedings{caron2021emerging,
  title={Emerging properties in self-supervised vision transformers},
  author={Caron, Mathilde and Touvron, Hugo and Misra, Ishan and J{\'e}gou, Herv{\'e} and Mairal, Julien and Bojanowski, Piotr and Joulin, Armand},
  booktitle={Proceedings of the IEEE/CVF International Conference on Computer Vision},
  pages={9650--9660},
  year={2021}
}

@inproceedings{adaloglou2023exploringlimitsdeepimage,
  author={Nikolas Adaloglou and Felix Michels and Hamza Kalisch and Markus Kollmann},
  title={Exploring the Limits of Deep Image Clustering using Pretrained Models},
  year={2023},
  cdate={1672531200000},
  pages={297-299},
  url={http://proceedings.bmvc2023.org/297/},
  booktitle={The British Machine Vision Conference}
}

@inproceedings{wassersteingan,
  title = 	 {{W}asserstein Generative Adversarial Networks},
  author =       {Martin Arjovsky and Soumith Chintala and L{\'e}on Bottou},
  booktitle = 	 {Proceedings of the 34th International Conference on Machine Learning},
  pages = 	 {214--223},
  year = 	 {2017},
  url = 	 {https://proceedings.mlr.press/v70/arjovsky17a.html},
}

@article{lazic2021,
  title={Optimization issues in {KL}-constrained approximate policy iteration},
  author={Lazi{\'c}, Nevena and Hao, Botao and Abbasi-Yadkori, Yasin and Schuurmans, Dale and Szepesv{\'a}ri, Csaba},
  journal={arXiv preprint arXiv:2102.06234},
  year={2021}
}

@inproceedings{khosla2020supervised,
  title={Supervised contrastive learning},
  author={Khosla, Prannay and Teterwak, Piotr and Wang, Chen and Sarna, Aaron and Tian, Yonglong and Isola, Phillip and Maschinot, Aaron and Liu, Ce and Krishnan, Dilip},
  booktitle={Advances in Neural Information Processing Systems},
  volume={33},
  pages={18661--18673},
  year={2020}
}

@inproceedings{he2016deep,
  title={Deep residual learning for image recognition},
  author={He, Kaiming and Zhang, Xiangyu and Ren, Shaoqing and Sun, Jian},
  booktitle={Proceedings of the IEEE Conference on Computer Vision and Pattern Recognition},
  pages={770--778},
  year={2016}
}

@techreport{krizhevsky2009learning,
  title={Learning multiple layers of features from tiny images},
  author={Krizhevsky, Alex},
  year={2009},
  url={https://www.cs.toronto.edu/~kriz/learning-features-2009-TR.pdf},
  institution={University of Toronto}
}

@inproceedings{deng2009imagenet,
  title={{ImageNet}: A large-scale hierarchical image database},
  author={Deng, Jia and Dong, Wei and Socher, Richard and Li, Li-Jia and Li, Kai and Fei-Fei, Li},
  booktitle={Proceedings of the IEEE Conference on Computer Vision and Pattern Recognition},
  pages={248--255},
  year={2009}
}

@article{van2008visualizing,
  title={Visualizing data using {t-SNE}},
  author={Maaten, Laurens van der and Hinton, Geoffrey},
  journal={Journal of Machine Learning Research},
  volume={9},
  pages={2579--2605},
  year={2008}
}

@article{hinton2002sne,
  title={Stochastic neighbor embedding},
  author={Hinton, Geoffrey E and Roweis, Sam},
  journal={Advances in Neural Information Processing Systems},
  volume={15},
  year={2002}
}

@book{rogers1964packing,
  title={Packing and Covering},
  author={Rogers, C Ambrose},
  volume={54},
  year={1964},
  publisher={University Press Cambridge}
}

@article{kingma2014adam,
  title={Adam: A method for stochastic optimization},
  author={Kingma, Diederik P},
  journal={International Conference on Learning Representations},
  year={2015}
}

\appendix

\newpage

\section{Supplementary Figures and Tables}\label{apd:figures_tables}

\begin{figure}[!h]
\centering
\includegraphics[width=1\textwidth]{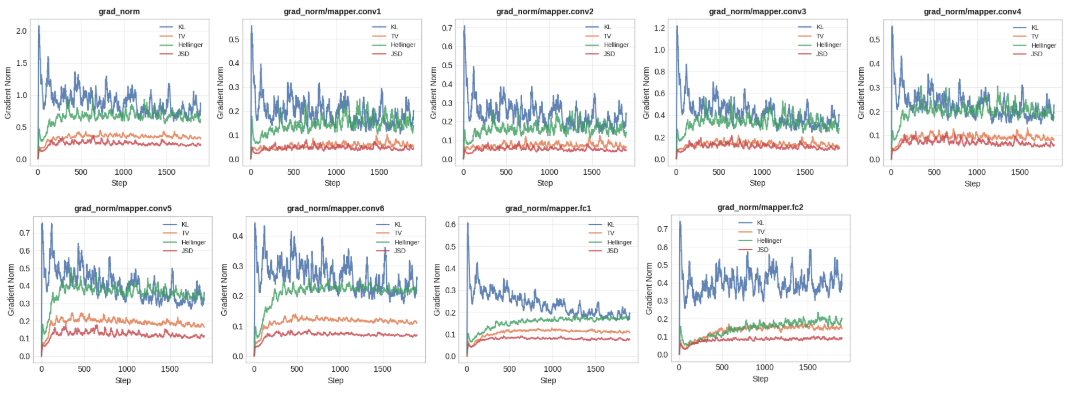}
\caption{Gradient norms for each divergence from running SNE on CIFAR-10 images with a CNN backbone. KL's unbounded nature creates initialization instability that manifests consistently across all network layers, while bounded divergences (TV, Hellinger, JSD) provide more stable gradient behavior throughout training.}
\label{fig:norm}
\end{figure}



\section{Geometric Analysis of the Crowding Problem}\label{apd:geometry}

In this appendix, we illustrate with an example why the KL divergence loss results in overcrowding in SNE.

Let the original high-dimensional space have $d$ dimensions.
A $d$-dimensional space permits $d + 1$ clusters/classes that are equidistant from each other --- 
say they're distance 1 from each other.
Now, mapping these clusters to a reduced $d'$-dimensional feature space ($d' \ll d$) while ensuring the clusters
are still separated by distance at least 1 means that some clusters will be very far from each other
($\Omega\big(d^{1/d'}\big)$) \citep{rogers1964packing}.
This incurs a high KL penalty as $q(j \mid i) \ll p(j \mid i)$
for data points $i$, $j$ from two clusters that are now much farther away each other.
Thus, minimizing the KL loss inevitably leads to some clusters being brought too close together (crowding)
in the low-dimensional feature space to prevent a high KL penalty from two clusters that are too far apart.
Bounded divergences like TV, Hellinger, and JSD resolve this issue since
$q(j \mid i) \ll p(j \mid i)$ incurs a lower penalty.

\end{document}